\documentclass[11pt,a4paper]{article}
\usepackage[hyperref]{eacl2021}

\usepackage{times}
\usepackage{latexsym}
\usepackage{hyperref}
\usepackage{verbatim}
\usepackage{graphicx} 

\usepackage{epigraph}
\usepackage{amsmath}
\usepackage{booktabs}
\usepackage{multirow}
\usepackage{float}
\usepackage{caption}
\usepackage{subfigure}
\usepackage{microtype}
\usepackage[ruled,vlined]{algorithm2e}

\usepackage{booktabs}

\usepackage{dblfloatfix}   

\aclfinalcopy 
\def\aclpaperid{1061} 

\title{`Just because you are right, doesn't mean I am wrong': \\ Overcoming a Bottleneck in the Development and Evaluation of \\ Open-Ended Visual Question Answering (VQA) Tasks}

\author{Man Luo, Shailaja Keyur Sampat, Riley Tallman, Yankai Zeng, \\ \textbf{Manuha Vancha, Akarshan Sajja,   Chitta Baral} \\
  Arizona State University, Tempe, AZ, USA \\
  \texttt{\{mluo26,ssampa17,rtallman,yzeng55,mvancha,asajja,chitta\}@asu.edu} \\}

\date{}

\begin{document}
\maketitle
\begin{abstract}


GQA~\citep{hudson2019gqa} is a dataset for real-world visual reasoning and compositional question answering. We found that many answers predicted by the best vision-language models on the GQA dataset do not match the ground-truth answer but still are semantically meaningful and correct in the given context. In fact, this is the case with most existing visual question answering (VQA) datasets where they assume only one ground-truth answer for each question. We propose Alternative Answer Sets (AAS) of ground-truth answers to address this limitation, which is created automatically using off-the-shelf NLP tools. We introduce a semantic metric based on AAS and modify top VQA solvers to support multiple plausible answers for a question. We implement this approach on the GQA dataset and show the performance improvements\footnote{Code and data are available in this link \url{https://github.com/luomancs/alternative_answer_set.git}}.

\end{abstract}

\section{Introduction}

One important style of visual question answering (VQA) task involves open-ended responses such as free-form answers or fill-in-the-blanks. The possibility of multiple correct answers and multi-word responses makes the evaluation of open-ended tasks harder, which has forced VQA datasets to restrict answers to be a single word or a short phrase. Despite enforcing these constraints, from our analysis of the GQA dataset~\citep{hudson2019gqa}, we noticed that a significant portion of the visual questions have issues. For example, a question \textit{ ``Who is holding the bat?"} has only one ground truth answer \textit{``batter"} while other reasonable answers like \textit{``batsman"}, \textit{``hitter"} are not credited. We identified six different types of issues with the dataset and illustrated them in Table \ref{gqa-problems}.

A large-scale human-study conducted by~\citep{gurari2017crowdverge} on VQA~\citep{antol2015vqa} and VizWiz~\citep{gurari2019vizwiz} found that almost 50\% questions in these datasets have multiple possible answers.  datasets had similar observations. The above evidence suggests that it is unfair to penalize models if their predicted answer is correct in a given context but does not match the ground truth answer.

With this motivation, we leverage existing knowledge bases and word embeddings to generate Alternative Answer Sets (AAS) instead of considering visual questions to have fixed responses. Since initially obtained AAS are generated from multiple sources and observed to be noisy, we use textual entailment to verify semantic viability of plausible answers to make alternative answer sets more robust. We justify the correctness and quality of the generated AAS by human evaluation. 
We introduce a semantic metric based on AAS and train two vision-language models LXMERT~\citep{tan2019LXMERT} and ViLBERT~\citep{lu2019vilbert} on two datasets. The experimental results show that the AAS metric evaluates models' performances more reasonably than the old metric. Lastly, we incorporate AAS in the training phase and show that it further improves on the proposed metric. Figure \ref{fig:process} gives an overview of our work. 
\begin{table*}[ht]

\begin{tabular}{@{}llc@{}}
\toprule
\multicolumn{1}{c}{\textbf{Issue Type}} & \multicolumn{1}{c}{\textbf{Definition}}                                                                                                                                   & \textbf{\%} \\ \midrule
{[}1{]} Synonym and Hypernym            & \begin{tabular}[c]{@{}l@{}}Synonym or hypernym of the ground-truth can also be \\ considered as a correct answer for a given question-image pair.\end{tabular}            & 9.1         \\
{[}2{]} Singular/Plural                 & \begin{tabular}[c]{@{}l@{}}Singular or plural of the ground-truth can also be considered \\ as a correct answer for a given question-image pair.\end{tabular}             & 1.0         \\
{[}3{]} Ambiguous Objects               & \begin{tabular}[c]{@{}l@{}}Question refers to an object but the image contains multiple\\ such objects that can lead to different possible answers.\end{tabular}          & 5.8         \\
{[}4{]} Multiple Correct Answers        & \begin{tabular}[c]{@{}l@{}}If a given image-question pair is not precise, annotators might \\ have different opinion which leads to multiple correct answers\end{tabular} & 7.0         \\
{[}5{]} Missing Object(s)               & Object referred in the question is not clearly visible in image.                                                                                                          & 4.3         \\
{[}6{]} Wrong Label                     & The ground-truth answer to a question-image pair is incorrect.                                                                                                            & 6.7         \\ \bottomrule
\end{tabular}
\caption{Six types of issues observed in the GQA dataset, their definition and their distribution observed in manual review of 600 samples from testdev balanced split. For example of each issue type, refer Figure \ref{fig:issue-examples}.}
\label{gqa-problems}
\end{table*}

\begin{figure*}[]

\centering
\subfigure{
    \begin{minipage}[t]{0.31\linewidth}
    \centering
    \small
    [1] Synonym and Hypernym
    \includegraphics[width=0.95\linewidth, height=2.5cm]{n317260}
    \tiny
    \begin{tabular}{l}
        {\bf Question:} Who is holding the bat? \\
        {\bf Ground-truth:} batter \\
        {\bf Explanation:} `batsman' is a synonym of 'batter' \\
        {\bf SU-AAS:} batter, batsman, hitter, ballplayer, player
        
    \end{tabular}
    \end{minipage}
}
\subfigure{
    \begin{minipage}[t]{0.31\linewidth}
    \centering
    \small
    [2] Singular/Plural
    \includegraphics[width=0.95\linewidth, height=2.5cm]{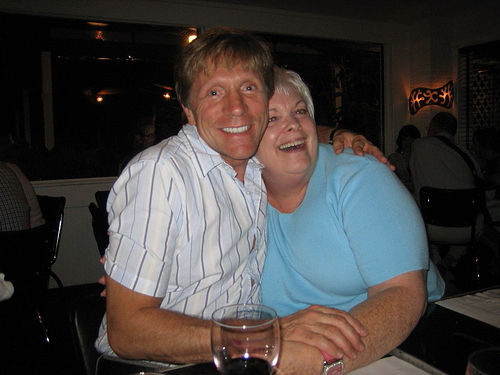}
    \tiny
    \begin{tabular}{l}
        {\bf Question:} Who is wearing the dress? \\
        {\bf Ground-truth:} women \\
        {\bf Explanation:} singular form `woman' is also correct\\ (there is only one woman in the picture) \\
        {\bf SU-AAS:} women, female, woman, people, adult female
    \end{tabular}
    \end{minipage}
}
\subfigure{
    \begin{minipage}[t]{0.31\linewidth}
    \centering
    \small
    [3] Ambiguous Objects  \\
    \includegraphics[width=0.95\linewidth, height=2.5cm]{n326988}
    \tiny
    \begin{tabular}{l}
        {\bf Question:} Does the person in front of the cabinets \\
         have brunette color? \\
        {\bf Ground-truth:} Yes \\
        {\bf Explanation:} there are two people in front of the cabinet \\ and it is not clear which person is being referred to \\
        {\bf SU-AAS:} yes
    \end{tabular}
    \end{minipage}
}
\subfigure{
    \begin{minipage}[t]{0.31\linewidth}
    \centering
    \small
    [4] Multiple Correct Answers 
    \includegraphics[width=0.95\linewidth, height=2.5cm]{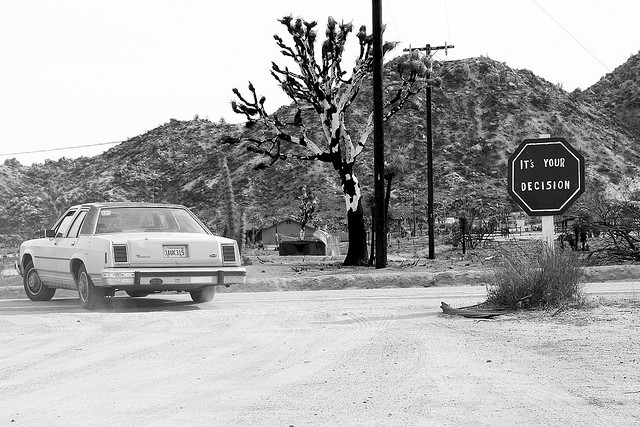}
    \tiny
    \begin{tabular}{l}
        {\bf Question:} Which place is it? \\
        {\bf Ground-truth:} road \\
        {\bf Explanation:} some person might answer `road' and \\ some might answer `street' \\
        {\bf SU-AAS:} road, street, roadway, paved
    \end{tabular}
    \end{minipage}
}
\subfigure{
    \begin{minipage}[t]{0.31\linewidth}
    \centering
    \small
    [5] Missing Object(s) 
    \includegraphics[width=0.95\linewidth, height=2.5cm]{n546616}
    \tiny
    \begin{tabular}{l}
        {\bf Question:} Does the marker look narrow? \\
        {\bf Ground-truth:} Yes \\
        {\bf Explanation:} the `marker' is missing from the image    \\
        {\bf SU-AAS:} yes
    \end{tabular}
    \end{minipage}
}
\subfigure{
    \begin{minipage}[t]{0.31\linewidth}
    \centering
    \small
    [6] Wrong Label 
    \includegraphics[width=0.95\linewidth, height=2.5cm]{n497658}
    \tiny
    \begin{tabular}{l}
        {\bf Question:} Do the door and the logo have the same color? \\
        {\bf Ground-truth:} Yes \\
        {\bf Explanation:} the correct answer is `no' as the door is \\ white and logo is green. \\
        {\bf SU-AAS:} yes
    \end{tabular}
    \end{minipage}
}

\caption{\small Examples from GQA dataset for each issue type and SU-AAS i.e. AAS of ground-truth based on semantic union approach. SU-AAS can resolve Synonym and Hypernym, Singular/Plural, and  Multiple Correct Answers for a given problem.}
\label{fig:issue-examples}
\end{figure*}

\section{Related Works}

We discuss related works from two aspects, dataset creation and evaluation.

\paragraph{Dataset Creation-Level}

Large-scale VQA datasets are often curated through crowd-sourcing, where open-ended ground-truths are determined by majority voting or annotator agreement. 
The subjectivity in crowd-sourced datasets is well-studied in human-computer interaction literature-~\citep{gurari2016visual,gurari2017crowdverge,yang2018visual} etc.~\citet{ray2018make} suggested creating a semantically-grounded set of questions for consistent answer predictions. 
\citep{bhattacharya2019does} analyzed VQA and VizWiz datasets to present 9-class taxonomy of visual questions that suffer from subjectivity and ambiguity.
Our analysis on GQA partially overlaps with this study. GQA dataset only provides one ground truth for each question; thus, we propose AAS to extend answers by phrases with close semantic meaning as the ground-truth answer.

\paragraph{Evaluation-Level}
For open-ended VQA tasks, the standard accuracy metric can be too stringent as it requires a predicted answer to exactly match the ground-truth answer. To deal with different interpretations of words and multiple correct answers,~\citet{malinowski2014multi} defined a WUPS scoring from lexical databases with Wu-Palmer similarity~\citep{10.3115/981732.981751}. ~\citet{abdelkarim2020long} proposed a soft matching metric based on wordNet~\citep{miller1998wordnet} and word2vec~\citep{mikolov2013distributed}. Different from them, we incorporate more advanced NLP resources tools to generate answer sets and rely on textural entailment to validate semantics for robustness. Semantic evaluation has also discussed for other tasks, such as image captioning generation ~\citep{feinglass2021smurf}.

\section{Analysis of GQA Dataset}\label{sec:analysis of dataset}

GQA is a dataset for real-world visual reasoning and compositional question answering. Instead of human annotation, answers to the questions in GQA are generated from the scene graphs of images. We found that automatic creation leads to flaws in the dataset; thus, we manually analyze 600 questions from the testdev balanced split of GQA dataset, and identify six issues shown in Table \ref{gqa-problems}. Figure \ref{fig:issue-examples} shows examples of each type of issue. 

These issues are caused by (not limited to) three reasons. First, the dataset assumes only one ground truth so that other answers with semantic closed meaning are ignored. We propose AAS to address this issue to some extent and describe AAS in the next section. Second, some questions referring to multiple objects cause ambiguous meaning. We leverage scene graphs to address this issue and found 2.92\% and 2.94\% ambiguous questions in balanced training split and balanced validation split, respectively. These ambiguous questions can be removed from the dataset. Third, there are incorrect scene graph detections so that some questions and/or labels do not match with the given images. We plan to address these issues in our future work.

\section{Alternative Answer Set}\label{sec:approach}

\begin{figure*}
  \center
  \includegraphics[width=\linewidth]{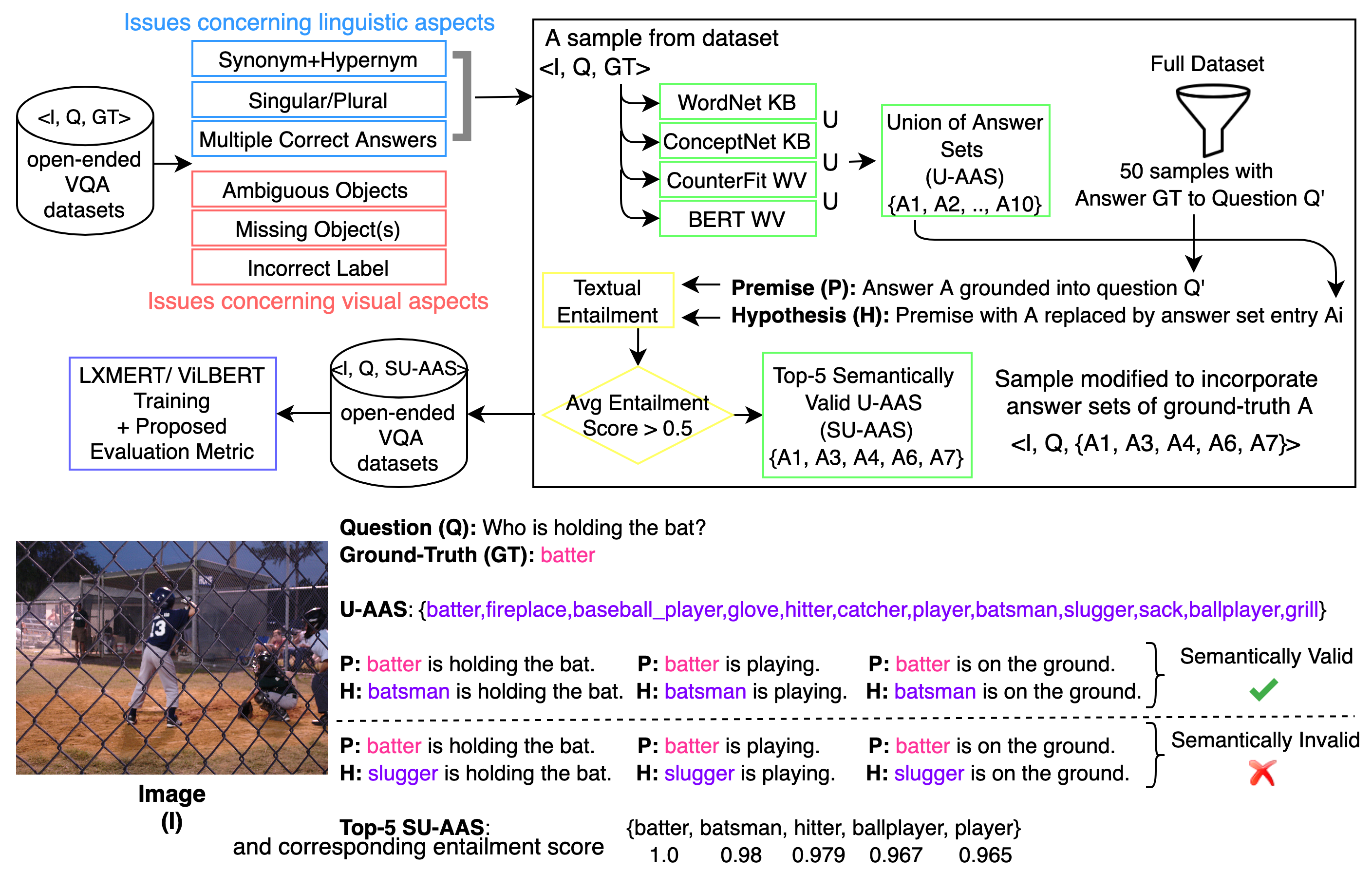}
  \caption{(top) The workflow for generating Alternative Answer Set (AAS) for VQA datasets (bottom) An example from GQA dataset showing semantically valid AAS for the answer `batter' generated using above workflow}
  \label{fig:process}
\end{figure*}

To credit answers with semantically close meaning as the ground-truth, we propose 
a workflow that can be visualized from Figure \ref{fig:process}. Each item in VQA dataset consists of $<$I, Q, GT$>$, where I is an image, Q is a question, and GT is a ground-truth answer. We define an Alternative Answer Set (AAS) as a collection of phrases  \{A$_1$, A$_2$, A$_3$,.., A$_n$\} such that A$_i$ replaced with GT is still a valid answer to the given Image-Question pair. We construct AAS for each unique ground-truth automatically from two knowledge bases: Wordnet~\citep{miller1998wordnet} and ConcpetNet~\citep{liu2004conceptnet}, two word embeddings: BERT ~\citep{devlin2018bert} and counter-fitting~\citep{mrkvsic2016counter}. We assign a semantic score to each alternative answer by textural entailment and introduce the AAS metric. 

\subsection {Semantic Union AAS}

We take a union of four methods to find all alternative answers. For example, ``stuffed animal" is semantic similar to ``teddy bear", which appears in the AAS based on BERT but not in WordNet. However, the union might include phrases that we want to distinguish from the label like ``man" is in the AAS of ``woman" when using the BERT-based approach. For this reason, we employ the textural entailment technique to compute a semantic score of each alternative answer. For each label, we first obtain 50 sentences containing the ground-truth label from GQA dataset. We take each sentence as a premise, replace the label in this sentence with a phrase in its AAS as a hypothesis to generate an entailment score between 0-1. Specifically, we use publicly available RoBERTa~\citep{liu2019roberta} model trained on SNLI (Stanford Natural Language Inference)~\citep{bowman-etal-2015-large} dataset for entailment computation. The semantic score of the alternative answer is the average of 50 entailment scores. If the semantic score is lower than the threshold of 0.5, then this alternative answer is thrown out. We choose 0.5 since it is the middle of 0 and 1. 

Lastly, we sort the AAS by semantic score and keep the top K in the semantic union AAS, annotated by SU-AAS. We experiment with different values of K from 2 to 10, and decide K to be 6, a trade-off between accuracy and robustness. Note that the performance of textual entailment model is a contributing factor in obtaining quality AAS. Therefore, we recommend using the state-of-the-art entailment model when our proposed method is applied on other VQA datasets. 

\subsection{Evaluation Metric Based on AAS}

We propose AAS metric and semantic score: given a question Q$_i$, an image I$_i$, the alternative answer set of GT$_i$ denoted by S$_{\text{GT}_i}$, the prediction of model P$_i$ is correct if and only if it is found in S$_{\text{GT}_i}$, and the score of P$_i$ is S$_{\text{GT}_i}$(P$_i$), where  S$_{\text{GT}_i}$(P$_i$) is the semantic score of P$_i$.  Mathematically, \begin{equation*}
\label{eq:eval}
\text{Acc(Q$_i$, I$_i$, S$_{\text{GT}_i}$, {P}$_i$)} = \begin{cases}
\text{S}_{\text{GT}_i}(\text{P}_i) & \text{if P}_i \in \text{S}_{\text{GT}_i}\\
0 &\text{else}
\end{cases}
\end{equation*}

\section{Experiments}

In this section, we first show that the performance of vision-language models on two datasets is improved based on the AAS metric. Then, we describe our experiment to incorporate AAS with one model on GQA dataset. Last, we verify the correctness of AAS by human evaluation. 

\subsection{Baseline Methods}
\begin{figure}[h]
\centering
\includegraphics[width=7cm]{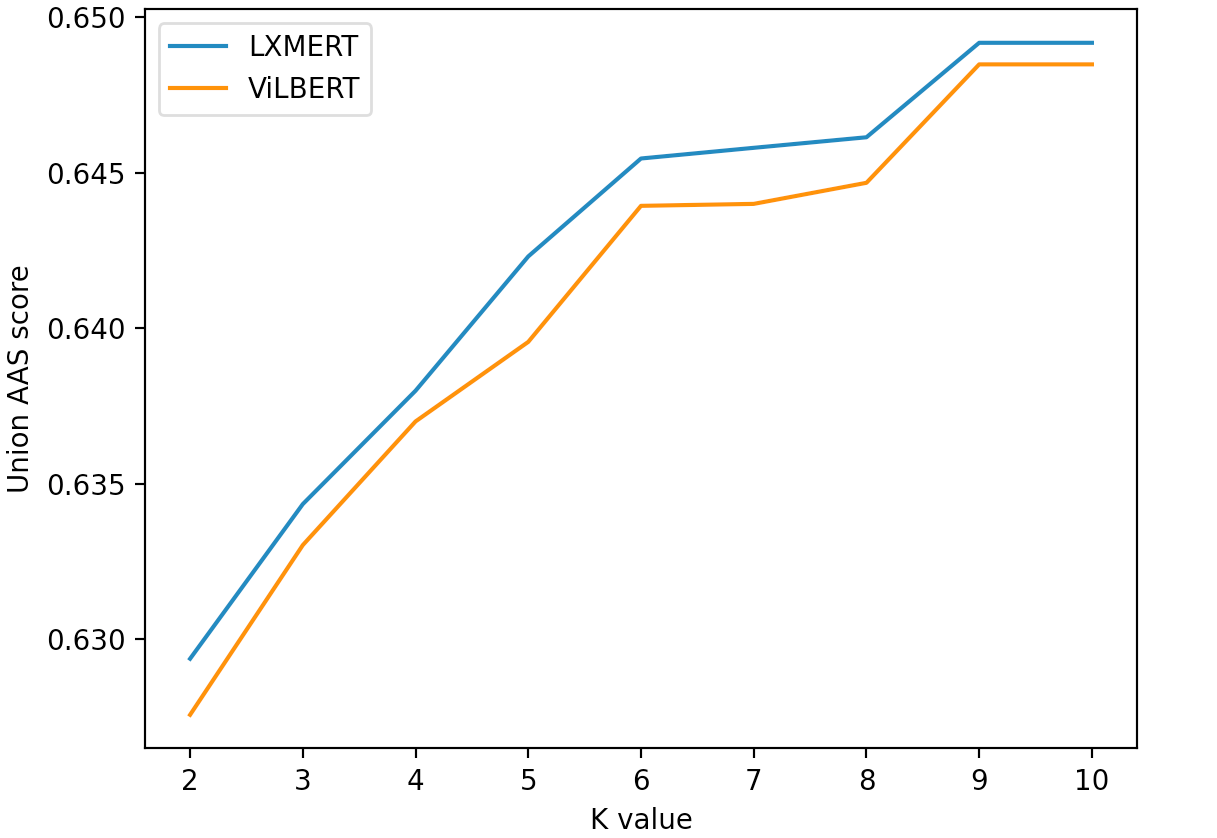}
\caption{Union AAS score of different value of K}
\label{fig:kvalue}
\end{figure}

We select two top Vision-and-Language models, ViLBERT~\citep{lu2019vilbert} and LXMERT ~\citep{tan2019LXMERT} and evaluate their performances based on the AAS metric. From Table \ref{tab:bentchmark}, we see that for the GQA dataset, LXMERT and ViLBERT have 4.49\%, 4.26\%  improvements on union AAS metric separately.  For VQA2.0 dataset, LXMERT and ViLBERT have 0.82\%, 0.53\%  improvements on union AAS metric separately. It is expected that the improvement on VQA2.0 dataset is less than GQA since the former dataset already provides multiple correct answers. Figure \ref{fig:kvalue} shows the impacts of the value K of Union AAS on the scores. From the figure, we see that when K increases from 2 to 6, the score gets increased significantly, and slightly when k increases from 6 to 9, but not increases more after K is 9. Since values 7 and 8 do not significantly improve the score, and the value 9 introduces noise, we take the top 6 as the SU-AAS. 

\begin{table*}[ht]
\centering
\small
\renewcommand{\arraystretch}{1.2}
\resizebox{0.9\textwidth}{!}{
\begin{tabular}{llcccccc}
\hline
Dataset & Model &  Original Metric & WordNet & BERT & CounterFit & ConceptNet & Union\\ \hline
GQA & LXMERT  & 60.06  & 61.79 &  62.69   & 62.75  &  63.58       & {\bf64.55}  \\ 
(testdev) & ViLBERT  & 60.13 & 61.90  &  62.69 & 62.74   &    63.67  &  {\bf 64.39}   \\ \hline

VQA   & LXMERT & 69.98  &    70.21   &  70.54 &    70.33        & 70.52  & {\bf 70.80}   \\
 (valid) & ViLBERT  & 77.65  &  77.82 &  78.10  &   77.93  &   78.06  &  {\bf 78.28} \\ \hline
\end{tabular}
}
\caption{The evaluation of two models on GQA and VQA with original metric and AAS based metrics. }
\label{tab:bentchmark}
\end{table*}

\vspace{-0.5cm}

\subsection{Training with AAS} 

We incorporate SU-AAS of ground truth in training phase, so the model learns that more than one answer for a given example can be correct. We train LXMERT on GQA dataset with this objective.

Table \ref{tab:newtraining} shows the results of LXMERT trained with AAS compared with the baseline. Not surprisingly, the performance evaluated on the original method drops because the model has a higher chance to predict answers in AAS, which are different from the ground truth, and thus the performance evaluated on SU-AAS metric increases. 
\begin{table}[H]
    \centering
    \renewcommand{\arraystretch}{1.2}
    \resizebox{0.48\textwidth}{!}{
    \begin{tabular}{lccccc}
    \hline
    \multirow{2}{*}{\bf Dataset}  & \multicolumn{2}{c}{\textbf{ Exact Matching Accuracy }} &  \multicolumn{2}{c}{\textbf{SU-AAS Accuracy }}\\
        &LXMERT & LXMERT$_{AAS}$   &LXMERT & LXMERT$_{AAS}$ \\\hline
    GQA(testdev) & {\bf 60.06} & 59.02     & 64.55  & {\bf 65.22} \\
    \hline
    \end{tabular}
    }
    
    \caption{\footnotesize Incorporate AAS in the training phase of LXMERT (LMXERT$_{AAS}$) on GQA dataset.}
    \label{tab:newtraining}
\end{table}

\subsection{Evaluation of AAS}

To validate the correctness of AAS, we measure the correlation between human judgment and AAS. Specifically, for each label of GQA, we take the SU-AAS and ask three annotators to justify if alternative answers in AAS can replace the label. If the majority of annotators agree upon, we keep the answer in the AAS, remove otherwise. In this way, we collect the human-annotated AAS. We compare the human-annotated AAS with each automatically generated AAS. We take the intersection over union (IoU) score to evaluate the correlation between automatic approach and human annotation: a higher IoU score means stronger alignment. 
\begin{table}[H]
\renewcommand{\arraystretch}{1.2}
\centering
\resizebox{0.5\textwidth}{!} {
    \begin{tabular}{lccccc}
        \hline
        \textbf{Method} & WordNet & BERT & CounterFit &ConceptNet & Union  \\
        \hline
        \textbf{IoU\%} &  48.25 &  56.18  & 58.95 & 58.39 & 80.5\\
        \hline
    \end{tabular}
}
\caption{\small The IoU scores between human annotations and AAS based on five approaches.}
\label{aas-iou}
\end{table} 

\section{Discussion and Conclusion} 

To evaluate a model from a semantic point of view, we define an alternative answer set (AAS). We develop a workflow to automatically create robust AAS for ground truth answers in the dataset using Textual Entailment. Additionally, we did human verification to assess the quality of automatically generated AAS. The high agreement score indicates that entailment model is doing a careful job of filtering relevant answers. From experiments on two models and two VQA datasets, we show the effectiveness of AAS-based evaluation using our proposed metric.  

AAS can be applied to other tasks, for example, machine translation. BLEU\citep{papineni2002bleu} score used to evaluate machine translation models incorporates an average of n-gram precision but does not consider the synonymy. 
Therefore, METEOR~\citep{banerjee2005meteor} was proposed to overcome this problem. However, METEOR only relies on the synset of WordNet to get the synonyms. Our proposed AAS 
has the advantage of both knowledge base and word embeddings, which would help better evaluate translation tasks.

\section*{Acknowledgments}
We are thankful to Tejas Gokhale for useful discussions and feedback on this work. We also thank anonymous reviewers for their thoughtful feedback. This work is partially supported by the National Science Foundation grant IIS-1816039.

\bibliography{eacl2021}
\bibliographystyle{acl_natbib}

\end{document}